\documentstyle[psfig,11pt]{article}
\newcommand{\eqpunc}[1]{{\makebox[0pt][l]{\qquad\rm{#1}}}} 
\newcommand{\mlog}[1]{\tilde{#1}}

\newcommand{\path}[3]{P_{#1}(#2,#3)}
\newcommand{\upath}[4]{P_{#1}^{#3}(#2,#4)}
\newcommand{\from}[2]{P_{#1}(#2)}
\newcommand{\ufrom}[3]{P_{#1}^{#3}(#2)}
\newcommand{\den}[1]{[\![ #1 ]\!]}
\newcommand{\pvar}[1]{{#1}}
\newcommand{\ajoin}{\bowtie}
\newcommand{\src}[1]{{#1}.\mbox{src}}
\newcommand{\dst}[1]{{#1}.\mbox{dst}}
\newcommand{\lab}[1]{{#1}.\mbox{lab}}
\newcommand{\cost}[1]{{#1}.\mbox{w}}
\newcommand{\fcost}[1]{F(#1)}
\newcommand{\proj}[2]{\pi_{#1}(#2)}
\newcommand{\fst}[1]{\proj{1}{#1}}
\newcommand{\snd}[1]{\proj{2}{#1}}
\newcommand{\inp}[1]{{#1}.\mbox{in}}
\newcommand{\outp}[1]{{#1}.\mbox{out}}
\newcommand{\Skip}[2]{\mbox{Skip}_{#1}(#2)}
\newcommand{\Mark}[2]{\mbox{Mark}_{#1}(#2)}

\def\argmax{\mathop{\rm argmax}}

\newcommand{\opt}{\max}
\newcommand{\argopt}{\argmax}
\newcommand{\freecost}{1}
\newcommand{\worstcost}{0}
\newcommand{\extend}{\times}
\newcommand{\collect}{+}
\newcommand{\bigcollect}{\sum}
\newcommand{\reals}{{\bf R}}

\title{Speech Recognition by Composition of Weighted Finite Automata}
\author{Fernando C.~N.~Pereira \and Michael D.~Riley \\
AT\&T Research \\
600 Mountain Ave., Murray Hill, NJ 07974}
\date{\today}
\begin{document}
\maketitle
\bibliographystyle{plain}
\begin{abstract}
We present a general framework based on weighted finite automata and weighted
finite-state transducers for describing and implementing speech
recognizers. The framework allows us to represent
uniformly the information sources and data structures used in
recognition, including context-dependent units, pronunciation
dictionaries, language models and lattices. Furthermore, general but
efficient algorithms can used for combining information sources in
actual recognizers and for optimizing their application.
In
particular, a single {\em composition} algorithm is used both to combine
in advance information sources such as language models and dictionaries,
and to combine acoustic observations and information sources dynamically
during recognition.
\end{abstract}
\section{Introduction}
Many problems in speech processing can be
usefully analyzed in terms of the ``noisy channel'' metaphor: given an
observation sequence $o$, find which intended message $w$ is most
likely to generate that observation sequence by maximizing 
\[P(w,o) = P(o|w) P(w),\]
where $P(o|w)$ characterizes the {\em transduction} between intended
messages and observations, and $P(w)$ characterizes the message
generator. More generally, the transduction between messages and
observations may involve several {\em stages} relating successive {\em
levels} of
representation:
\begin{equation}
\begin{array}{r@{=}l}
P(s_0,s_k) & P(s_k|s_0) P(s_0) \\
P(s_k|s_0) & \sum_{s_1,\ldots,s_{k-1}} P(s_k|s_{k-1}) \cdots
P(s_{1}|s_0)
\end{array} \label{sum-cascade} 
\end{equation}
Each $s_j$ is a sequence of units of an appropriate representation,
for instance phones or syllables in
speech recognition.
A straightforward but useful observation is that any such a cascade
can be factored at any intermediate level
\begin{equation}
P(s_j|s_i) = \sum_{s_l} P(s_j|s_l) P(s_l|s_i) \label{cascade-factor}
\end{equation}

For computational reasons, sums and products in (\ref{sum-cascade})
are often replaced by minimizations and sums of negative log
probabilities, yielding the approximation
\begin{equation}
\begin{array}{rcl}
\mlog{P}(s_0,s_k) & = & \mlog{P}(s_k|s_0) + \mlog{P}(s_0) \\
\mlog{P}(s_k | s_0) &\approx & \min_{s_1,\ldots,s_{k-1}} \sum_{1\le j\le
k} \mlog{P}(s_j|s_{j-1}) \end{array}
 \label{min-cascade}
\end{equation}
where $\mlog{X}=-\log X$.
In this formulation, assuming the approximation is reasonable,
the most likely message $s_0$ is the one minimizing
$\mlog{P}(s_0,s_k)$.

In current speech recognition systems, a transduction stage is
typically modeled by a finite-state device, for example a hidden
Markov model (HMM). However, the commonalities among stages are
typically not exploited, and each stage is represented and implemented
by ``ad hoc'' means. The goal of this paper is to show that the theory
of weighted rational languages and transductions can be used as a
general framework for transduction cascades. Levels of representation
will be modeled as weighted languages, and transduction stages will be
modeled as weighted transductions.

This foundation provides a rich set of operators for combining cascade
levels and stages that generalizes the standard operations on regular
languages, suggests novel ways of combining models of different parts
of the decoding process, and supports uniform algorithms for
transduction and search throughout the cascade.  Computationally,
stages and levels of representation are represented as weighted finite
automata, and a general automata {\em composition} algorithm
implements the relational composition of successive stages. Automata
compositions can be searched with standard best-path algorithms to
find the most likely transcriptions of spoken utterances. A ``lazy''
implementation of composition allows search and pruning to be carried
out concurrently with composition so that only the useful portions of
the composition of the observations with the decoding cascade is
explicitly created. Finally, finite-state minimization techniques can
be used to reduce the size of cascade levels and thus improve
recognition efficiency \cite{Mohri:seqtrans}.

Weighted languages and transductions are generalizations of the
standard notions of language and transduction in formal language
theory \cite{Berstel:79,Harrison:intro}. A weighted
language is a mapping from strings over an alphabet to weights, while
a weighted transduction is a mapping from pairs of strings over two
alphabets to weights. For example, when weights represent
probabilities and assuming appropriate normalization, a weighted
language is just a probability distribution over strings, and a
weighted transduction a conditional probability distribution between
strings. The weighted {\em rational} languages and transducers are
those that can be represented by weighted finite-state acceptors
(WFSAs) and weighted finite-state transducers (WFSTs), as described in
more detail in the next section.  In this paper we will be concerned
with the weighted rational case, although some of the theory can be
profitably extended more general language classes closed under
intersection with regular languages and composition with rational
transductions \cite{Lang:89,Teitelbaum:73}.

The notion of weighted rational transduction arises from the
combination of two ideas in automata theory: rational transductions,
used in many aspects of formal language theory \cite{Berstel:79}, and
weighted languages and automata, developed in pattern recognition
\cite{Booth+Thompson:73,Paz:71} and algebraic automata theory
\cite{Berstel+Reutenauer:88,Eilenberg:74,Kuich+Salomaa:86}.  Ordinary
(unweighted) rational transductions have been successfully applied by
researchers at Xerox PARC \cite{Kaplan+Kay:94} and at the University
of Paris 7
\cite{Mohri:compact,Mohri:local-grammars,Roche:93,Silberztein:dict},
among others, to several problems in language processing, including
morphological analysis, dictionary compression and syntactic
analysis. HMMs and probabilistic finite-state language models can be
shown to be equivalent to WFSAs. In algebraic automata theory,
rational series and rational transductions \cite{Kuich+Salomaa:86} are
the algebraic counterparts of WFSAs and WFSTs and give the correct
generalizations to the weighted case of the standard algebraic
operations on formal languages and transductions, such as union,
concatenation, intersection, restriction and composition. We believe
our work is the first application of these generalizations to speech
processing.

While we concentrate here on speech recognition applications, the same
framework and tools have also been applied to other language
processing tasks such as the segmentation of Chinese text into words
\cite{Sproat+al:segmentation}.  We explain how a standard HMM-based
recognizer can be naturally viewed as equivalent to a cascade of
weighted transductions, and how the approach requires no modification
to accommodate context dependencies that cross higher-level unit
boundaries, for instance cross-word context-dependent models.  This is
an important advantage of the transduction approach over the usual,
but more limited ``substitution'' approach used in existing to speech
recognizers. Substitution replaces a symbol at a higher level by its
defining language at a lower level, but, as we will argue, cannot
model directly the interactions between context-dependent units at the
lower level.

\section{Theory}
\subsection{The Weight Semiring}
As discussed informally in the previous section, our approach relies
on associating {\em weights} to the strings in a language, the string
pairs in a transduction and the transitions in an automaton.  The
operations used for weight combination should reflect the intended
interpretation of the weights. For instance, if the weights of
automata transitions represent transition probabilities, the weight
assigned to a path should be the product of the weights of its
transitions, while the weight (total probability) assigned to a set of
paths with common source and destination should be the sum of the
weights of the paths in the set. However, if the weights represent
negative log-probabilities and we are operating under the Viterbi
approximation that replaces the sum of the probabilities of
alternative paths by the probability of the most probable path, path
weights should be the sum of the weights of the transitions in the
path and the weight assigned to a set of paths should be the minimum
of the weights of the paths in the set. Both of these weight
structures are special cases of {\em commutative semirings}, which are
the basis of the general theory of weighted languages, transductions
and automata
\cite{Berstel+Reutenauer:88,Eilenberg:74,Kuich+Salomaa:86}.

In general, a {\em semiring} is a set $K$ with two binary operations,
{\em collection} $\collect_K$ and {\em extension} $\extend_K$, such that:
\begin{itemize}
\item collection is associative and commutative with identity $\worstcost_K$;
\item extension is associative with identity $\freecost_K$;
\item extension distributes over collection;
\item $a\extend_K\worstcost_K=\worstcost_K\extend_K a=\worstcost$ for any $a\in K$.
\end{itemize}
The semiring is {\em commutative} if extension is commutative.

Setting $K=\reals^+$ with $+$ for collection, $\times$ for extension,
0 for $\worstcost_K$ and 1 for $\freecost_K$ we obtain the {\em
sum-times} semiring, which we can use to model probability
calculations.  Setting $K=\reals^+\cup\{ \infty \}$ with $\min$ for
collection, $+$ for extension, $\infty$ for $\worstcost_K$ and 0 for
$\freecost_K$ we obtain the {\em min-sum} semiring, which models
negative log-probabilities under the Viterbi approximation.

In general, weights represent some measure of ``goodness'' that we
want to optimize. For instance, with probabilities we are interested
in the highest weight, while the lowest weight is sought for
negative log-probabilities. We thus assume a total
order on weights and write $\opt_x f(x)$ for the optimal value of the
weight-valued function $f$ and $\argopt_x f(x)$ for some $x$ that
optimizes $f(x)$. We also assume that extension and collection are
monotonic with respect to the total order.

In what follows, we will assume a fixed semiring $K$ and thus drop the
subscript $K$ in the symbols for its operations and identity
elements. Unless stated otherwise, all the discussion will apply to
any commutative semiring, if necessary with a total order for
optimization. Some definitions and calculations involve collecting
over potentially infinite sets, for instance the set of strings of a
language. Clearly, collecting over an infinite set is always
well-defined for {\em idempotent} semirings such as the min-sum
semiring, in which $a\collect a=a\;\forall a\in K$. More generally, a
{\em closed} semiring is one in which collecting over infinite sets is
well defined. Finally, some particular cases arising in the discussion
below can be shown to be well defined for the plus-times semiring
under certain mild conditions on the weights assigned to strings or
automata transitions \cite{Booth+Thompson:73,Kuich+Salomaa:86}.

\subsection{Weighted Transductions and Languages} 
In the transduction cascade (\ref{sum-cascade}), each stage
corresponds to a mapping from input-output pairs $(r,s)$ to
probabilities $P(s|r)$.  More formally, stages in the cascade will be
{\em weighted transductions\/} $T: \Sigma^{*}\times
\Gamma^{*}\rightarrow K$ where $\Sigma^{*}$ and $\Gamma^{*}$ are the
sets of strings over the alphabets $\Sigma$ and $\Gamma$, and $K$ is
the weight semiring.  We will denote by $T^{-1}$ the {\em inverse\/}
of $T$ defined by $T(t,s) = T(s,t)$.

The right-most step of (\ref{sum-cascade}) is not a transduction, but
rather an information source, the language model. We will represent
such sources as {\em weighted languages} $L:\Sigma^{*}\rightarrow K$.

Each transduction $S: \Sigma^{*}\times \Gamma^{*}\rightarrow K$ has
two associated weighted languages, its its {\em first} and {\em second
projections\/} $\fst{S}:\Sigma^{*}\rightarrow K$ and
$\snd{S}:\Gamma^{*}\rightarrow K$, defined by
\[
\begin{array}{lcr}
\fst{S}(s) & = &\bigcollect_{t\in \Gamma^{*}}S(s,t) \\
\snd{S}(t) & = &\bigcollect_{s\in \Sigma^{*}}S(s,t)
\end{array}
\]

Given two transductions $S: \Sigma^{*}\times \Gamma^{*}\rightarrow K$
and $T: \Gamma^{*}\times \Delta^{*}\rightarrow K$, we define their
{\em composition} $S\circ T$ by
\begin{equation}
(S\circ T)(r,t) = \bigcollect_{s\in \Gamma^{*}} S(r,s)\extend T(s, t)\label{compose}
\end{equation}
For example, if 
$S$ represents $P(s_l|s_i)$ and $T$ $P(s_j|s_l)$ in (\ref{cascade-factor}),
$S\circ T$ represents $P(s_j|s_i)$.

A weighted transduction $S: \Sigma^{*}\times
\Gamma^{*}\rightarrow K$ can be also {\em applied} to a weighted language
$L:\Sigma^{*}\rightarrow K$ to yield
a weighted language $S[L]$ over $\Gamma$:
\begin{equation}
S[L](s) = \bigcollect_{r\in \Sigma^{*}} L(r) \extend S(r, s)\label{apply}
\end{equation}

We can also identify any weighted language $L$ with the identity
transduction restricted to $L$:
\[
L(r,r') = \left\{\begin{array}{ll}L(r) & \mbox{\rm if $r=r'$} \\
                                 \worstcost & \mbox{\rm otherwise}
                \end{array}\right.
\]
Using this identification, application is transduction
composition followed by projection:
\[\begin{array}{lcr}
\snd{L\circ S}(s) & = &\bigcollect_{r\in \Sigma^{*}}\bigcollect_{r'\in \Sigma^{*}} L(r,r')\extend
S(r', s) \\
& = & \bigcollect_{r\in \Sigma^{*}}L(r,r)\extend S(r, s) \\
& = & \bigcollect_{r\in \Sigma^{*}}L(r)\extend S(r, s) \\
& = & S[L](s)
\end{array}
\]
From now on, we will take advantage of the identification of languages
with transductions and use $\circ$ to express both composition and
application, often leaving implicit the projections required to
extract languages from transductions.  In particular, the {\em
intersection} of two weighted languages $M, N: \Sigma^{*}\rightarrow
K$ is given by
\begin{equation}
\fst{M\circ N}(s) = \snd{M\circ N}(s)  = M(s) \extend N(s)\label{intersect}
\end{equation}

It is easy to see that composition is associative, that is, the result of
any transduction cascade $R_1\circ\cdots\circ R_m$ is independent of
order of application of the composition operators.
\begin{figure}
\centerline{\psfig{figure=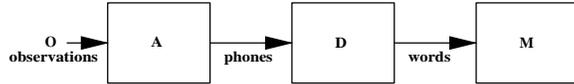,width=3in}}
\caption{Recognition Cascade}
\label{recog}
\end{figure}

\begin{table*}
\begin{center}
\begin{tabular}{l|l|l}
& Transduction \\
\hline
singleton & $\{(u,v)\}(w,z) = \freecost$ iff
$u=w$ and $v=z$ \\
scaling & $(kT)(u,v) = k \extend T(u,v)$ \\
sum & $(S + T)(u,v) =
S(u,v) \collect T(u,v)$ \\
concatenation & $(ST)(t,w) =
\bigcollect_{rs=t,uv=w} S(r,u) \extend T(s,v)$ \\
power &
 $\begin{array}[t]{lcl}T^{0}(\epsilon,\epsilon) & = &\freecost \\
                            T^{0}(u\ne \epsilon,v\ne \epsilon) & = & \worstcost \\
                            T^{n+1} & = & T T^{n} \end{array}$ \\
closure & $T^{*} = \bigcollect_{k\ge0} T^{k}$
\end{tabular}
\end{center}
\caption{Rational Operations}
\label{ops}
\end{table*}

For a more concrete example, consider the transduction cascade for
speech recognition depicted in Figure \ref{recog}, where $A$ is the
transduction from acoustic observation sequences to phone sequences,
$D$ the transduction from phone sequences to word sequences
(essentially a pronunciation dictionary) and $M$ a weighted language
representing the language model. Given a particular sequence of
observations $o$, we can represent it as the trivial weighted language
$O$ that assigns $1$ to $o$ and $0$ to any other sequence. Then
$O\circ A$ represents the acoustic likelihoods of possible phone
sequences that generate $o$, $O\circ A\circ D$ the acoustic-lexical
likelihoods of possible word sequences yielding $o$, and $ O\circ
A\circ D\circ M$ the combined acoustic-lexical-linguistic
probabilities of word sequences generating $o$.  The word string $w$
with the highest weight in  $\snd{O\circ A\circ D\circ M}$ is the
most likely sentence hypothesis generating $o$.

Composition is thus the main operation involved in the construction
and use of transduction cascades. As we will see in the next section,
composition can be implemented as a suitable generalization of the
usual intersection algorithm for finite automata.  In addition to
composition, weighted transductions (and languages, given the
identification of languages with transductions presented earlier) can
be constructed from simpler ones using the operations shown in Table
\ref{ops}, which generalize in a straightforward way the regular
operations well-known from traditional automata theory
\cite{Harrison:intro}. In fact, the rational languages and
transductions are exactly those that can be built from singletons by
applications of scaling, sum, concatenation and closure.

For example, assume that for each word $w$ in a lexicon we are given a
rational transduction $D_w$ such that $D_w(p,w)$ is the probability
that $w$ is realized as the phone sequence $p$. Note that this
allows for multiple pronunciations for $w$. Then the
rational transduction $\left(\sum_w D_w\right)^{*}$ gives the
probabilities for realizations of word sequences as phone sequences if
we leave aside cross-word context dependencies, which will be
discussed in Section \ref{sec:sr}.

\subsection{Weighted Automata}

Kleene's theorem states that regular languages are exactly those
representable by finite-state acceptors
\cite{Harrison:intro}. Generalized to the weighted case and to
transductions, it states that weighted rational languages and
transductions are exactly those that can be represented by weighted
finite automata \cite{Eilenberg:74,Kuich+Salomaa:86}.  Furthermore,
all the operations on languages and transductions we have discussed
have finite-automata counterparts, which we have implemented. Any
cascade representable in terms of those operations can thus be
implemented directly as an appropriate combination of the programs
implementing each of the operations.

A {\em $K$-weighted finite automaton} $A$ is given by a finite set of states
$Q_A$, a set of {\em transition labels} $\Lambda_A$, an initial state
$i_A$, a {\em final weight} function $F_A:Q_A\rightarrow K$,
\footnote{The usual notion of final state can be represented by
$F_{A}(q) = \freecost $ if $q$ is final, $F_{A}(q) = \worstcost$
otherwise.  More generally, we call a state {\em final} if its weight
is not $\worstcost$. Also, we will interpret any non-weighted automaton
as a weighted automaton in which all transitions and final states
have weight $\freecost$.} and a finite set $\delta_{A} \subset Q_A
\times \Lambda_A \times K \times Q_A$ of {\em transitions}
$t=(\src{t},\lab{t},\cost{t},\dst{t})$.  The label set $\Lambda_A$
must have with an associative {\em concatenation} operation $u\cdot v$
with identity element $\epsilon_{A}$. A {\em weighted finite-state
acceptor} (WFSA) is a $K$-weighted finite automaton with
$\Lambda_{A}=\Sigma^{*}$ for some finite alphabet $\Sigma$. A {\em
weighted finite-state transducer} (WFST) is a $K$-weighted finite
automaton such that $\Lambda_{A}=\Sigma^{*}\times\Gamma^{*}$ for given
finite alphabets $\Sigma$ and $\Gamma$, its label concatenation is
defined by $(r,s)\cdot(u,v) = (ru,sv)$, and its identity (null) label
is $(\epsilon,\epsilon)$. For $l=(r,s)\in \Sigma^{*}\times\Gamma^{*}$
we define $\inp{l} = r$ and $\outp{l} = s$.  As we have done for
languages, we will often identify a weighted acceptor with the
transducer with the same state set and a transition $(q,(x,x),k,q')$
for each transition $(q,x,k,q')$ in the acceptor.

A {\em path} in an automaton $A$ is a sequence of transitions 
$\pvar{p}=t_1,\ldots, t_m$ in $\delta_A$ with $\src{t_i}=\dst{t_{i-1}}$
for $1 < i \le k$. We define the {\em source} and the {\em
destination} of $\pvar{p}$ by $\src{\pvar{p}}=\src{t_1}$ and
$\dst{\pvar{p}} = \dst{t_m}$, respectively.
\footnote{For convenience, for each state $q\in Q_A$ we also have an
{\em empty path} with no transitions and source and destination $q$.}
The {\em label} of $\pvar{p}$ is the concatenation $\lab{\pvar{p}}
=\lab{t_1}\cdot \cdots \cdot \lab{t_m}$, its {\em weight} is the
product $\cost{\pvar{p}} = \cost{t_1}\extend \cdots \extend
\cost{t_m}$ and its {\em acceptance weight} is $\fcost{\pvar{p}}=
\cost{\pvar{p}}\extend F_{A}(\dst{\pvar{p}})$. We denote by
$\path{A}{q}{q'}$ the set of all paths in $A$ with source $q$ and
destination $q'$, by $\from{A}{q}$ the set of all paths in $A$ with
source $q$, by $\upath{A}{q}{u}{q'}$ the subset of $\path{A}{q}{q'}$
with label $u$ and by $\ufrom{A}{q}{u}$ the subset of $\from{A}{q}$
with label $u$.

Each state $q\in Q_A$ defines a weighted transduction (or a weighted
language):
\begin{equation}\label{eq:statelang}
L_A(q)(u) = \bigcollect_{\pvar{p}\in \ufrom{A}{q}{u}}\fcost{\pvar{p}}\eqpunc{.}
\end{equation}
Finally, we can define the weighted transduction (language) of a
weighted transducer (acceptor) $A$ by
\begin{equation}\label{eq:autlang}
\den{{A}} = L_A(i_A)\eqpunc{.}
\end{equation}
The appropriate generalization of Kleene's
theorem to weighted acceptors and transducers states that under
suitable conditions guaranteeing that the inner sum in
(\ref{eq:statelang}) is defined, 
weighted rational languages and transductions are
exactly those defined by weighted automata as outlined here
\cite{Kuich+Salomaa:86}.

Weighted acceptors and transducers are thus faithful implementations
of rational languages and transductions, and all the operations on
these described above have corresponding implementations in terms of
algorithms on automata. In particular,
composition is implemented by the automata operation we
now describe.

\subsection{Automata Composition}
\label{sec:composition}
Informally, the composition of two automata $A$ and $B$ is a generalization
of NFA intersection. Each state in the composition is a pair of a
state of $A$
and a state of $B$, and each path in the composition corresponds to a pair of
a path in $A$ and a path in $B$ with compatible labels. The total weight of
the composition path is the extension of the weights of
the corresponding paths in $A$ and $B$. The composition operation thus
formalizes the notion of coordinated search in two graphs, where the
coordination corresponds to a suitable agreement between path labels.

The more formal discussion that follows will be presented in terms of
transducers, taking advantage the identifications of languages
with transductions and of acceptors with transducers given earlier.

Consider two transducers ${A}$ and ${B}$ with $\Lambda_{A}
= \Sigma^{*} \times \Gamma^{*}$ and $\Lambda_{B}
= \Gamma^{*} \times \Delta^{*}$. Their composition ${A}\ajoin{B}$
will be a transducer with $\Lambda_{{A}\ajoin {B}} =
\Sigma^{*} \times \Delta^{*}$ such that:
\begin{equation}
\label{eq:compositionspec}
\den{{A}\ajoin {B}} = \den{{A}} \circ \den{B}\eqpunc{.}
\end{equation}
By definition of $L_{\cdot}(\cdot)$ and $\circ$ we have for any $q\in Q_A$
and $q'\in Q_B$:
\begin{equation}
\begin{array}{rcl}
\lefteqn{(L_A(q) \circ L_B(q'))(u,w)} \\
& = & \bigcollect_{v\in\Gamma^{*}}
(\bigcollect_{\pvar{p}\in \ufrom{A}{q}{(u,v)}} \fcost{\pvar{p}})\extend
(\bigcollect_{\pvar{p}'\in \ufrom{B}{q'}{(v,w)}} \fcost{\pvar{p}'}) \\ & = &
\bigcollect_{v\in\Gamma^{*}} \bigcollect_{\pvar{p}\in
\ufrom{A}{q}{(u,v)}}\bigcollect_{\pvar{p}'\in \ufrom{B}{q'}{(v,w)}}
\fcost{\pvar{p}}\extend \fcost{\pvar{p}'} \\
& = & \bigcollect_{(\pvar{p},\pvar{p}')\in J(q,q',u,w)}
\fcost{\pvar{p}} \extend \fcost{\pvar{p}'}
\label{eq:compositionpath}
\end{array}
\end{equation}
where
$J(q,q',u,w)$ is the set of pairs $(p,p')$ of paths $\pvar{p} \in
\from{A}{q}$ and $\pvar{p}'\in\from{B}{q'}$ such that
$\inp{\lab{\pvar{p}}} = u$, $\outp{\lab{\pvar{p}}}=
\inp{\lab{\pvar{p}'}}$ and 
$\outp{\lab{\pvar{p}'}} = w$.
In particular, we have:
\begin{equation}
(\den{A} \circ \den{B})(u,w) = \bigcollect_{(\pvar{p},\pvar{p}')\in
J(i_A,i_B,u,w)}\fcost{\pvar{p}} \extend \fcost{\pvar{p}'}
\eqpunc{.}
\end{equation}
Therefore, assuming that (\ref{eq:compositionspec}) is satisfied, this
equation collects the weights of all paths $\pvar{p}$ in
${A}$ and $\pvar{p}'$ in ${B}$ such that $\pvar{p}$ maps $u$
to some string $v$ and $\pvar{p}'$ maps $v$ to $w$. In particular, on
the min-sum weight semiring, the shortest path labeled $(u,w)$
in $\den{{A}\ajoin {B}}$ minimizes the sum of the costs of
paths labeled $(u,v)$ in ${A}$ and $(v,w)$ in ${B}$, for
some $s$. 

We will give first the construction of $A\ajoin B$ for $\epsilon$-free
transducers $A$ and $B$, that is, those with transition labels in
$\Sigma\times\Gamma$ and $\Gamma\times\Delta$, respectively. Then
${A}\ajoin{B}$ has state set $Q_{{A}\ajoin{B}} = Q_{A} \times Q_{B}$,
initial state $i_{{A}\ajoin{B}}= (i_{A},i_{B})$ and final weights
$F_{{A}\ajoin{B}}(q,q') = F_{A}(q)F_{B}(q')$. Furthermore, there is a
transition $((q,q'),(x,z),k\extend k',(r,r')) \in \delta_{{A}\ajoin{B}}$ iff
there are transitions $(q,(x,y),k,r)\in\delta_{A}$ and
$(q',(y,z),k',r')\in\delta_{B}$. This construction is similar to
the standard intersection construction for DFAs; a proof that it
indeed implements transduction composition (\ref{eq:compositionspec}) is given
in Appendix \ref{sec:proofs}.

\begin{figure}
\centerline{\psfig{figure=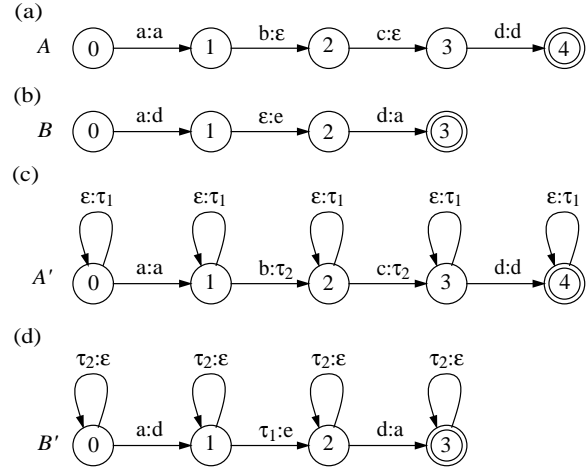,width=3in}}
\caption{Transducers with $\epsilon$ Labels}
\label{fig:ab}
\end{figure}

\begin{figure}
\centerline{\psfig{figure=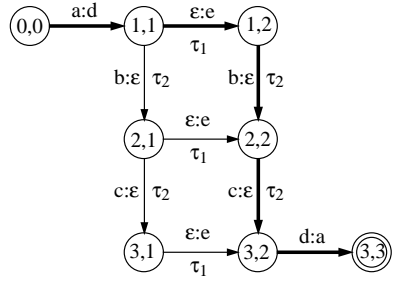,width=2in}}
\caption{Composition with Marked $\epsilon$s}
\label{fig:naivecomposition}
\end{figure}

\begin{figure}
\centerline{\psfig{figure=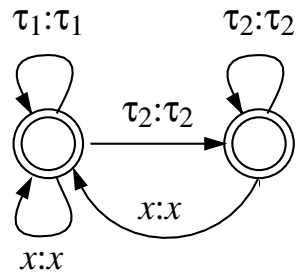,width=1.5in}}
\caption{Filter Transducer}
\label{fig:filter}
\end{figure}

In the general case, we consider transducers $A$ and $B$
with labels over $\Sigma^{?}\times\Gamma^{?}$ and
$\Gamma^{?}\times\Delta^{?}$, respectively, where $\Lambda^{?} =
\Lambda \cup \{\epsilon\}$. \footnote{It is easy to see that any
transducer with transition labels in $\Sigma^{*}\times\Gamma^{*}$ is
equivalent to a transducer with labels in
$\Gamma^{?}\times\Delta^{?}$.} 
As shown in (\ref{eq:compositionpath}), the composition
of $A$ and $B$ should have exactly one path for each pair of paths
$\pvar{p}$ in $A$ and $\pvar{p}'$ in $B$ with
\begin{equation}\label{eq:eqproj}
v = \outp{\lab{\pvar{p}}}=
\inp{\lab{\pvar{p}'}}\eqpunc{.}
\end{equation}
for some string $v \in \Gamma^*$ that we will call the {\em composition
string}. In the $\epsilon$-free case, it is clear that
$\pvar{p}=t_1,\ldots,t_m$, $\pvar{p}'= t'_1,\ldots,t'_m$ for some $m$
and $\outp{\lab{t_i}} = \inp{\lab{t'_i}}$. The pairing of $t_i$ with
$t'_i$ is precisely what the $\epsilon$-free composition construction
provides. In the general case, however, two paths $\pvar{p}$ and $\pvar{p}'$
satisfying (\ref{eq:eqproj}) need not have the same number of
transitions. Furthermore, there may be several ways to
align $\epsilon$ outputs in $A$ and $\epsilon$ inputs in $B$ with
staying in the same state in the opposite transducer. This is
exemplified by transducers $A$ and $B$ in Figure \ref{fig:ab}(a-b),
and the corresponding na\"{\i}ve composition in Figure \ref{fig:naivecomposition}. The
multiple paths from state $(1,1)$ to state $(3,2)$ correspond to different
interleavings between taking the transition from 1 to 2 in $B$ and the
transitions from 1 to 2 and from 2 to 3 in $A$. In the weighted case,
including all those paths in the composition would in general lead to an
incorrect total weight for the transduction of string $abcd$ to string
$da$. Therefore, we need a method for selecting a single composition path for
each pair of compatible paths in the composed transducer. 

The following construction, justified in Appendix
\ref{sec:gen-composition}, achieves the desired result. For label $l$,
define $\proj{1}{l} = \inp{l}$ and $\proj{2}{l} = \outp{l}$.  Given a
transducer $T$, compute $\Mark{i}{T}$ from $T$ by replacing the label
of every transition $t$ such that $\proj{i}{\lab{t}}=\epsilon$ with
the new label $l$ defined by $\proj{2-i}{l}=\proj{2-i}{\lab{t}}$ and
$\proj{i}{l}=\tau_i$, where $\tau_i$ is a new symbol. In words, each
$\epsilon$ on the $i$th component of a transition label is replaced by
$\tau_i$. Corresponding to $\epsilon$ transitions on one side of the
composition we need to stay in the same state on the other
side. Therefore, we define the operation $\Skip{i}{T}$ that for each
state $q$ of $T$ adds a new transition $(q,l,1,q)$ where
$\proj{2-i}{l}=\tau_i$ and $\proj{i}{l}=\epsilon$. We also need the
auxiliary transducer Filter shown in Figure \ref{fig:filter}, where
the transition labeled $x:x$ is shorthand for a set of transitions
mapping $x$ to itself (at no cost) for each $x\in\Gamma$.  Then for
arbitrary transducers $A$ and $B$, we have
\[
\den{A} \circ \den{B} = \den{\Skip{1}{\Mark{2}{A}}\ajoin\mbox{Filter}\ajoin
\Skip{2}{\Mark{1}{B}}} \eqpunc{.}
\]
For example, with respect to Figure \ref{fig:ab} we have
$A'=\Skip{1}{\Mark{2}{A}}$ and $B'=\Skip{2}{\Mark{1}{B}}$. The thick
path in Figure \ref{fig:naivecomposition} is the only one allowed by the
filter transduction, as desired.  In practice, the substitutions and
insertions of $\tau_i$ symbols performed by $\mbox{Mark}_i$ and
$\mbox{Skip}_i$ do not need to be performed explicitly, because
the effects of those operations can be computed on the fly by a
suitable implementation of composition with filtering.

The filter we described is the simplest to explain. In practice,
somewhat more complex filters, which we will describe elsewhere, help
reduce the size of the resulting transducer. For example, the filter
presented includes in the composition in states (2,1) and (3,1) on
Figure \ref{fig:naivecomposition}, from which no final state can be
reached. Such ``dead end'' paths can be a source of inefficiency when
using the results of composition.

\section{Speech Recognition}
\label{sec:sr}
\begin{figure}
\centerline{\psfig{figure=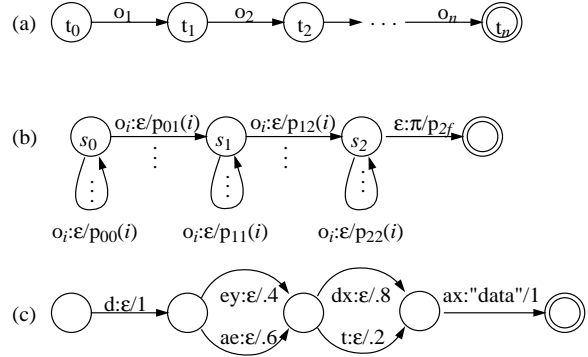,width=3in}}
\caption{Models as Automata}
\label{data}
\end{figure}

We now describe how to represent a speech recognizer as a composition
of transducers. Recall that we model the recognition task as the
composition of a language $O$ of acoustic observation sequences, a
transduction $A$ from acoustic observation sequences to phone
sequences, a transduction $D$ from phone sequences to word sequences
and a weighted language $M$ specifying the language model (see Figure
\ref{recog}). Each of these can be represented as a finite-state
automaton (to some approximation), denoted by the same name as the
corresponding transduction in what follows.

The acoustic observation automaton $O$ for 
a given utterance has the form shown on Figure \ref{data}a. 
Each state represents a fixed point in time $t_i$, and each
transition has a label, $o_i$, drawn from a finite alphabet 
that quantizes the acoustic signal between adjacent time
points and is assigned probability 1. \footnote{For more complex
acoustic distributions (for instance, continuous densities) we can
instead use multiple transitions $(t_{i-1},d,p(o_i|d),t_i)$ where $d$ is
an observation distribution and $p(o_i|d)$ the corresponding
observation probability.}

The transducer $A$ from acoustic observation sequences to phone
sequences is built from {\em phone models}. A phone model is a
transducer from sequences of acoustic observation labels to a specific
phone that assigns to each acoustic observation sequence the
likelihood that the specified phone produced it. Thus, different paths
through a phone model correspond to different acoustic realizations of
the phone.  Figure \ref{data}b shows a common topology for phone
models.  $A$ is then defined as the closure of the sum of the phone
models.

The transducer $D$ from phone sequences to word sequences is is built
similarly to $A$.  A {\em word model} is a transducer from phone
sequences to the specified word that assigns to each phone sequence
the likelihood that the specified word produced it. Thus, different
paths through a word model correspond to different phonetic
realizations of the word.  Figure \ref{data}c shows a typical topology
for a word model.  $D$ is then defined as the closure of the sum of
the word models.

Finally, the acceptor $M$ encodes the language model, for instance an
$n$-gram model. Combining those automata, we obtain $\proj{2}{O\ajoin
A\ajoin D\ajoin M}$, which assigns a probability to each word
sequence. The highest-probability path through that automaton
estimates the most likely word sequence for the given utterance.

The finite-state model of speech recognition that we have just
described is hardly novel. In fact, it is equivalent to that presented
in \cite{Bahl+al:83}, in the sense that it generates the same weighted
language. However, the transduction cascade approach presented here
allows one to view the computations in new ways.

For instance, because composition is associative, the computation of
$\argopt_w \proj{2}{O\ajoin A\ajoin D\ajoin M}(w)$ can be organized in
a variety of ways. In a traditional integrated-search recognizer, a
single large transducer is built in advance by $R= A\ajoin D\ajoin M$,
and used in recognition to compute $argmax_w \proj{2}{O \ajoin R}(w)$
for each observation sequence $O$ \cite{Bahl+al:83}. This approach is
not practical if the size of $R$ exceeds available memory, as is
typically the case for large-vocabulary speech recognition with
$n$-gram language models for $n > 2$. In those cases, pruning may be
interleaved with composition to to compute (an approximation of)
$((O\ajoin A) \ajoin D)\ajoin M$. Acoustic observations are first
transduced into a phone lattice represented as an automaton labeled by
phones (phone recognition). The whole lattice typically too big, so
the computation includes a pruning mechanism that generates only those
states and transitions that appear in high-probability paths.  This
lattice is in turn transduced into a word lattice (word recognition),
again possibly with pruning, which is then composed with the language
model \cite{Ljolje+Riley:92,Ljolje+al:stt-eurospeech}. The best
approach depends on the specific task, which determines the size of
intermediate results. By having a general package to manipulate
weighted automata, we have been able to experiment with various
alternatives.

So far, our presentation has used context-independent phone models. In
other words, the likelihood assigned by a phone model in $A$ is
assumed conditionally independent of neighboring phones. Similarly,
the pronunciation of each word in $D$ is assumed independent of
neighboring words. Therefore, each of the transducers has a
particularly simple form, that of the closure of the sum of (inverse)
{\em substitutions}. That is, each symbol in a string on the output
side replaces a language on the input side.  This replacement of a
symbol from one alphabet (for example, a word) by the automaton that
represents its substituted language from a over a finer-grained
alphabet (for example, phones) is the usual stage-combination
operation for speech recognizers \cite{Bahl+al:83}. 

However, it has been shown that context-dependent phone models, which
model a phone in the context of its adjacent phones, provide
substantial improvements in recognition accuracy
\cite{Lee:90}. Further, the pronunciation of a word will be affected
by its neighboring words, inducing context dependencies across word
boundaries.

We could include context-dependent models, such as triphone models, in
our presentation by expanding our `atomic models' in $A$ to one for
every phone in a distinct triphonic context. Each model will have the
same form as in Figure \ref{data}b, but it will be over an enlarged
output alphabet and have different likelihoods for the different
contexts. We could also try to directly specify $D$ in terms of the
new units, but this is problematic. First, even if each word in $D$
had only one phonetic realization, we could not directly substitute
its the phones in the realization by their context-dependent models,
because the given word may appear in the context of many different
words, with different phones abutting the given word. This problem is
commonly alleviated by either using left (right) context-independent
units at the word starts (ends), which decreases the model accuracy,
or by building a fully context-dependent lexicon and using special
machinery in the recognizer to insure the correct models are used at
word junctures. In either case, we can no longer use compact lexical
entries with multiple pronunciations such as that of Figure
\ref{data}c.
\begin{figure}
\centering
\mbox{\psfig{figure=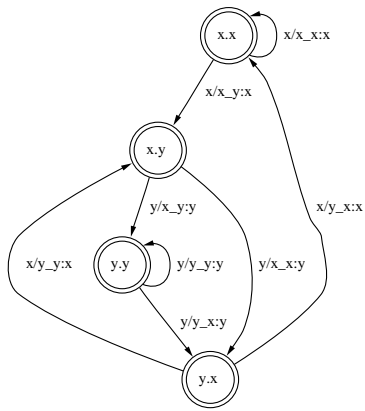,width=2in}}
\caption{Context-Dependency Transducer}
\label{context-dep}
\end{figure}
Those approaches attempt to solve the context-dependency problem by
introducing new substitutions, but substitutions are not really
appropriate for the task.

In contrast, context dependency can be readily represented by a simple
transducer. We leave $D$ as defined before, but interpose a new
transducer $C$ between $A$ and $D$ that convert between
context-dependent and context-independent units, that is, we now
compute $\argopt_w \proj{2}{O\ajoin A\ajoin C \ajoin D\ajoin M}(w)$. A
possible form for $C$ is shown in Figure \ref{context-dep}.  For
simplicity, we show only the portion of the transducer concerning two
hypothetical phones $x$ and $y$. The transducer maps each
context-dependent model $p/l\underline{\mbox{\hspace{1ex}}}r$,
associated to phone $p$ when preceded by $l$ and followed by $r$, to
an occurrence of $p$ which is guaranteed to be preceded by $l$ and
followed by $r$. To ensure this, each state labeled $p.q$ represents
the context information that all incoming transitions correspond to
phone $p$, and all outgoing transitions correspond to phone $q$.  Thus
we can represent context-dependency directly as a transducer, without
needing specialized context-dependency code in the recognizer. More
complex forms of context dependency such as those based on
classification trees over a bounded neighborhood of the target phone
can too be compiled into appropriate transducers and interposed in the
recognition cascade without changing any aspect of the recognition
algorithm. Transducer determinization and minimization techniques
\cite{Mohri:seqtrans} can be used to make context-dependency
transducers as compact as possible.

\section{Implementation}
The transducer operations described in this paper, together with a
variety of support functions, have been implemented in C. Two
interfaces are provided: a library of functions operating on an
abstract finite-state machine datatype, and a set of composable shell
commands for fast prototyping.  The modular organization of the
library and shell commands follows directly from their foundation in
the algebra of rational operations, and allows us to build new
application-specific recognizers automatically.

The size of composed automata and the efficiency of composition have
been the main issues in developing the implementation.  As explained
earlier, our main applications involve finding the highest-probability
path in composed automata. It is in general not practical to compute
the whole composition and then find the highest-probability path,
because in the worst case the number of transitions in a composition
grows with the product of the numbers of transitions in the composed
automata.  Instead, we have developed a lazy implementation of
composition, in which the states and arcs of the composed automaton
are created by pairing states and arcs in the composition arguments
only as they are required by some other operation, such as search, on
the composed automaton \cite{Riley+al:on-the-fly}. The use of an
abstract datatype for automata facilitates this, since functions
operating on automata do not need to distinguish between concrete and
lazy automata.

The efficiency of composition depends crucially on the efficiency with
which transitions leaving the two components of a state pair are
matched to yield transitions in the composed automaton. This task is
analogous to doing a relational join, and some of the sorting and
indexing techniques used for joins are relevant here, especially for
very large alphabets such as the words in large-vocabulary
recognition. The interface of the automaton datatype has been
carefully designed to allow for efficient transition matching while hiding
the details of transition indexing and sorting.

\section{Applications}

We have used our implementation in a variety of speech recognition and
language processing tasks, including continuous speech recognition in
the 60,000-word ARPA North American Business News (NAB) task
\cite{Ljolje+al:stt-eurospeech} and the 2,000-word ARPA ATIS task,
isolated word recognition for directory lookup tasks, and segmentation
of Chinese text into words \cite{Sproat+al:segmentation}.

The NAB task is by far the largest one we have attempted so far. In
our 1994 experiments \cite{Ljolje+al:stt-eurospeech}, we used a
60,000-word vocabulary, and several very large automata, including a
phone-to-syllable transducer with $5\times 10^{5}$ transitions, a
syllable-to-word (dictionary) transducer with $10^{5}$ transitions and
a language model (5-gram) with $3.4 \times 10^7$ transitions.  We are
at present experimenting with various improvements in modeling and in
the implementation of composition, especially in the filter, that
would allow us to use directly the lazy composition of the whole
decoding cascade for this application in a standard time-synchronous
Viterbi decoder. In our 1994 experiments, however, we had to break the
cascade into a succession of stages, each generating a pruned lattice
(an acyclic acceptor) through a combination of lazy composition and
graph search.  In addition, relatively simple models are used first
(context-independent phone models, bigram language model) to produce a
relatively small pruned word lattice, which is then intersected with
the composition of the full models to create a rescored lattice which
is then searched for the best path. That is, we use an approximate
word lattice to limit the size of the composition with the full
language and phonemic models. This multi-pass decoder achieved around
10\% word-error rate in the main 1994 NAB test, while requiring around
500 times real-time for recognition.

In our more recent experiments with lazy composition in synchronous
Viterbi decoders, we have been able to show that lazy composition is
as fast or faster than traditional methods requiring full expansion of
the composed automaton in advance, while requiring a small fraction of
the space.  The ARPA ATIS task, for example, uses a context transducer
with 40,386 transitions, a the dictionary with 4,816 transitions a
class-based variable-length $n$-gram language model
\cite{Riccardi+al:var-class-ngram} with 359,532 transitions. The
composition of these three automata would have around $6\times 10^6$
transitions. However, for a typical sentence only around 5\% of those
transitions are actually visited \cite{Riley+al:on-the-fly}.

\section{Further Work}
We have been investigating a variety of improvements, extensions and
applications of the present work. With Emerald Chung, we have been
refining the connection between a time-synchronous Viterbi decoder and
lazy composition to improve time and space efficiency. With Mehryar
Mohri, we have been developing improved composition filters, as well
as exploring on-the-fly and local determinization techniques for
transducers and weighted automata \cite{Mohri:seqtrans} to decrease
the impact of nondeterminism on the size (and thus the time required
to create) composed automata. Our work on the implementation has also
been influenced by applications to the compilation of weighted
phonological and morphological rules and by ongoing research on
integrating speech recognition with natural-language analysis and
translation. Finally, we are investigating applications to local
grammatical analysis, in which transducers have been often used but
not with weights.

\section*{Acknowledgments}
Hiyan Alshawi, Adam Buchsbaum, Emerald Chung, Don Hindle, Andrej
Ljolje, Mehryar Mohri, Steven Phillips and Richard Sproat have
commented extensively on these ideas, tested many versions of our
tools, and contributed a variety of improvements. Our joint work and
their own separate contributions in this area will be presented
elsewhere. The language model for the ATIS task was kindly supplied by
Enrico Bocchieri, Roberto Pieraccini and Giuseppe Riccardi. We would
also like to thank Raffaele Giancarlo, Isabelle Guyon, Carsten Lund
and Yoram Singer as well as the editors of this volume for many
helpful comments.

\bibliography{fsm}

\appendix

\section{Correctness of $\epsilon$-Free Composition}
\label{sec:proofs}
As shown in Section \ref{sec:composition} (\ref{eq:compositionpath}), we have
\begin{equation}
(L_A(q) \circ L_B(q'))(r,t) = 
\bigcollect_{s\in\Gamma^{*}} \bigcollect_{\pvar{p}\in
\ufrom{A}{q}{(r,s)}}\bigcollect_{\pvar{p}'\in \ufrom{B}{q'}{(s,t)}}
\fcost{\pvar{p}}\extend \fcost{\pvar{p}'}\eqpunc{.}\label{eq:compositionpath2}
\end{equation}
Clearly, for $\epsilon$-free transducers the variables $r, s, t,
\pvar{p}$ and $\pvar{p}'$ in this equation satisfy the constraint
$|r|=|s|=|t|=|\pvar{p}|=|\pvar{p}'|=n$ for some $n$. This allows us to show
the correctness of the composition construction for $\epsilon$-free automata
by induction on $n$. Specifically, we shall show that for any $q\in Q_A$
and $q'\in Q_B$
\begin{equation}\label{eq:corrcomposition}
L_{A\ajoin B}(q,q') = L_A(q) \circ L_B(q') \qquad .
\end{equation}

For $n=0$, from (\ref{eq:compositionpath2}) and the composition construction
we obtain
\[
\begin{array}{rcl}
(L_A(q)\circ L_B(q'))(\epsilon,\epsilon) & = & F_{A}(q)\extend F_{B}(q') \\
 & = & F_{A\ajoin B}(q,q') \\
& = & F_{A\ajoin B}(\epsilon,\epsilon)\qquad
\end{array}
\]
as needed. 

Assume now that $L_{A\ajoin B}(m,m')(u,w)=(L_A(m)\circ L_B(m'))(u,w)$
for any $m\in Q_A$, $m' \in Q_B$, $u\in \Sigma^{*}$ and $w\in
\Delta^{*}$ with $|u|=|w|<n$. Let $r=xu$ and $t=zw$, with $x\in\Sigma$
and $z\in\Delta$. Then by (\ref{eq:compositionpath2}) and the composition
construction we have
\[\begin{array}{rcll}
\lefteqn{(L_A(p) \circ L_B(q))(xu,zw)} \\ & = & \bigcollect_{y\in\Gamma}
\bigcollect_{v\in\Gamma^{*}} \bigcollect_{\pvar{p}\in \ufrom{A}{q}{(xu,yv)}}
\bigcollect_{\pvar{p}'\in \ufrom{B}{q'}{(yv,zw)}} \fcost{\pvar{p}}\extend
\fcost{\pvar{p}'} \\ & = & \begin{array}[t]{ll}\lefteqn{\bigcollect_{(q,(x,y),k,m)\in\delta_A}
\bigcollect_{(q',(y,z),k',m')\in\delta_B}} \\
& k\extend k'\extend (\bigcollect_{v\in\Gamma^{*}}\bigcollect_{\pvar{l}\in \ufrom{A}{m}{(u,v)}}
\bigcollect_{\pvar{l}'\in \ufrom{B}{m'}{(v,w)}} \fcost{\pvar{l}}\extend
\fcost{\pvar{l}'}) \end{array} \\ & = &
\begin{array}[t]{ll}\lefteqn{\bigcollect_{((q,q'),(x,z),j,(m,m'))\in\delta_{A\ajoin B}}} \\
&  j \extend (
\bigcollect_{v\in\Gamma^{*}}\bigcollect_{\pvar{l}\in \ufrom{A}{m}{(u,v)}}
\bigcollect_{\pvar{l}'\in \ufrom{B}{m'}{(v,w)}} \fcost{\pvar{l}}\extend
\fcost{\pvar{l}'}) \end{array}\\ & = &
\bigcollect_{((q,q'),(x,z),j,(m,m'))\in\delta_{A\ajoin B}} j\extend (L_A(m)\circ
L_B(m'))(u,w) \\ & = & \bigcollect_{((q,q'),(x,z),j,(m,m'))\in\delta_{A\ajoin
B}} j\extend  L_{A\ajoin B}(m,m')(u,w) \\ & = &
\bigcollect_{((q,q'),(x,z),j,(m,m'))\in\delta_{A\ajoin B}} j\extend 
(\bigcollect_{\pvar{g}\in \ufrom{A\ajoin B}{m,m'}{(u,w)}} W_{A\ajoin
B}(\pvar{g})) \\ & = & \bigcollect_{\pvar{h}\in \ufrom{A\ajoin
B}{q,q'}{(xu,zw)}} W_{A\ajoin B}(\pvar{h}) \\ & = & L_{A\ajoin
B}(q,q')(xu,zw) \qquad .
\end{array}
\]
This shows (\ref{eq:corrcomposition}) for $\epsilon$-free transducers, and as
a particular case 
\[
\den{{A}\ajoin {B}} = \den{{A}} \circ \den{B}\eqpunc{,}
\]
which states that transducer
composition correctly implements transduction composition.

\section{General Composition Construction}
\label{sec:gen-composition}

For any transition $t$ in $A$ or $B$, we define
\[\Mark{i}{t} = \left\{\begin{array}{ll} \tau_i & \mbox{if}\;
\proj{i}{\lab{t}}=\epsilon \\ \proj{i}{\lab{t}} &\mbox{otherwise}
\end{array}\right.\eqpunc{,}
\]
where each $\tau_i$ is a new symbol not in $\Gamma$.
This can be extended to a path $\pvar{p}=t_1,\ldots,t_m$ in the
obvious way by $\Mark{i}{\pvar{p}} =
\Mark{i}{t_1}\cdots\Mark{i}{t_m}$.  If $\pvar{p}$ and $\pvar{p}'$
satisfy (\ref{eq:eqproj}), there will be $m,n \ge k$ such that
$\pvar{p}=t_1,\ldots,t_m$, $\pvar{p}'=t'_1,\ldots,t'_n$, $v=y_1\cdots
y_k$ and $v = \outp{\lab{\pvar{p}}}= \inp{\lab{\pvar{p}'}}$. Therefore,
we will have $\Mark{2}{\pvar{p}} = u_0 y_1 u_1 \cdots u_{k-1} y_k u_k$
where $u_i\in \{\tau_2\}^*$ and $|u_0 \cdots u_k| = m-k$, and
$\Mark{1}{\pvar{p}'} = v_0 y_1 v_1 \cdots v_{k-1} y_k v_k$ where
$v_i\in \{\tau_1\}^*$ and $|v_0 \cdots v_k| = n-k$.

We will need the following standard definition of the {\em shuffle} $s
\star s'$ of two languages $L,L'\subseteq \Gamma^*$:
\[L \star L'= \{u_1v_1\cdots u_lv_l | u_1 \cdots u_l \in L, v_1 \cdots
v_l \in L'\}\eqpunc{.}
\]
Then it is easy to see that (\ref{eq:eqproj}) holds iff
\begin{equation}
J = (\{\Mark{2}{\pvar{p}}\}\star \{\tau_1\}^*) \cap (\{\Mark{1}{\pvar{p}'}\}\star
\{\tau_2\}^*) \neq \emptyset\eqpunc{.}\label{eq:taucompositionset}
\end{equation}
Each composition string $v\in J$ has the form 
\begin{equation}
v = v_0 y_1 v_1\cdots v_{k-1} y_k
v_k\label{eq:taucomposition}
\end{equation}
for $y_i\in \Gamma$ and $v_i\in \{\tau_1,\tau_2\}^*$.  Furthermore, by
construction, any string $v'_0 y_1 v'_1\cdots v'_{k-1} y_k v'_k$, where
each $v'_i$ is derived from $v_i$ by commuting $\tau_1$ instances with
$\tau_2$ instances, is also in $J$.

Consider for example the transducers $A$ shown in Figure \ref{fig:ab}a
and $B$ shown in Figure \ref{fig:ab}b. 
For path $\pvar{p}$ from state 0 to state 4 in $A$
and path $\pvar{p}'$ from state 0 to state 3 in $B$ we have the
following equalities:
\[
\begin{array}{rcl}
\Mark{2}{\pvar{p}} & = &a\tau_2\tau_2 d \\
\Mark{1}{\pvar{p}'} & = &a\tau_1 d \\
(\{\Mark{2}{\pvar{p}}\} \star \{\tau_1\}^*) \cap (\{\Mark{1}{\pvar{p}'}\} \star
\{\tau_2\}^*) & = & \left\{\begin{array}{l}a \tau_1 \tau_2 \tau_2 d, \\a \tau_2 \tau_1 \tau_2 d, \\
a \tau_2 \tau_2 \tau_1 d\end{array}\right\}
\end{array}
\]
Therefore, $\pvar{p}$ and $\pvar{p}'$ satisfy (\ref{eq:eqproj}), allowing
$\den{A}\circ \den{B}$ to map $abcd$ to $dea$. It is also
straightforward to see that, given the transducers $A'$ in Figure
\ref{fig:ab}c and $B'$ in Figure \ref{fig:ab}d, we have
\[
\begin{array}{rcl}
\{\Mark{2}{\pvar{p}}\}\star \{\tau_1\}^* & = &
\{\outp{\lab{\pvar{p}}} | \pvar{p}\in\from{A'}{0} \} \\
\{\Mark{1}{\pvar{p}'}\}\star \{\tau_2\}^* & = &
\{\inp{\lab{\pvar{p}'}} | \pvar{p}'\in\from{B'}{0} \} 
\end{array}
\]
Since there are no $\epsilon$ labels on the output side of $A'$ or the
input side of $B'$, we can apply to them the $\epsilon$-free composition
construction, with the result shown in Figure \ref{fig:naivecomposition}. Each of
the paths from the initial state to the final state corresponds to a
different composition string in $\{\Mark{2}{\pvar{p}}\}\star \{\tau_1\}^* \cap
\{\Mark{1}{\pvar{p}'}\}\star \{\tau_2\}^*$.

The transducer $A'\ajoin B'$ pairs up exactly the strings it should,
but it does not correctly implement $\den{A}\circ \den{B}$ in the
general weighted case. The construction described so far allows
several paths in $A'\ajoin B'$ corresponding to each pair of paths
from $A$ and $B$. Intuitively, this is possible because $\tau_1$ and
$\tau_2$ are allowed to commute freely in the composition string. But if one
pair of paths $\pvar{p}$ from $A$ and $\pvar{p}'$ from $B$ leads to
several paths in $A'\ajoin B'$, the weights from the
$\epsilon$-transitions in $A$ and $B$ will appear multiple times in
the overall weight for going from $(\src{\pvar{p}},\src{\pvar{p}'})$
to $(\dst{\pvar{p}},\dst{\pvar{p}'})$ in $A'\ajoin B'$. If the
semiring sum operation is not idempotent, that leads to the wrong
weights in (\ref{eq:compositionpath}).

To achieve the correct path multiplicity, we interpose a transducer
Filter between $A'$ and $B'$ in a 3-way composition
$\ajoin(A',\mbox{Filter},B')$. The Filter transducer is shown in
Figure \ref{fig:filter}, where the transition labeled $x:x$ represents
a set of transitions mapping $x$ to itself for each $x\in\Gamma$.  The
effect of Filter is to block any paths in $A'\ajoin B'$ corresponding
to a composition string containing the substring $\tau_2\tau_1$. This
eliminates all the composition strings (\ref{eq:taucomposition}) in
(\ref{eq:taucompositionset}) except for the one with $v_i\in
\{\tau_1\}^*\{\tau_2\}^*$, which is guaranteed to exist since $J$ in
(\ref{eq:taucompositionset}) allows all interleavings of $\tau_1$ and
$\tau_2$, including the required one in which all $\tau_2$ instances
must follow all $\tau_1$ instances. For example, Filter would remove
all but the thick-lines path in Figure \ref{fig:naivecomposition}, as needed
to avoid incorrect path multiplicities.

\end{document}